\documentclass[letterpaper, 10 pt, conference]{ieeeconf}
\IEEEoverridecommandlockouts
\usepackage{amsmath,amsfonts,amssymb}
\usepackage{array}
\usepackage[caption=false,font=normalsize,labelfont=sf,textfont=sf]{subfig}
\usepackage{textcomp}
\usepackage{stfloats}
\usepackage{url}
\usepackage{verbatim}
\usepackage{graphicx}
\usepackage{cite}
\usepackage{multirow}
\usepackage{placeins, afterpage}
\usepackage{cuted}
\usepackage{xcolor}
\usepackage{multicol}
\usepackage{mathtools}
\usepackage{lipsum}
\usepackage{float}
\usepackage{wrapfig}
\usepackage{orcidlink}
\usepackage{booktabs}
\usepackage{algorithm}
\usepackage{algpseudocode}
\usepackage{mdframed}
\usepackage[inkscapeformat=png]{svg}

\newcommand{\smallshift}{\;\;\;\;\;\;\;}

\DeclareMathOperator*{\argmax}{arg\,max}
\DeclareMathOperator*{\argmin}{arg\,min}
\newcommand{\edit}[1]{#1}

\begin{document}

\title{A Sliding-Window Filter for Online Continuous-Time \\ Continuum Robot State Estimation}

\author{\edit{Spencer Teetaert${}^1$, Sven Lilge${}^{1,2}$, Jessica Burgner-Kahrs${}^1$, Timothy D. Barfoot${}^1$
\thanks{This work was supported in part by the National Sciences and Engineering Research Council of Canada (NSERC) and the Ontario Graduate Scholarship.}% <-this % stops a space
\thanks{${}^1$ The authors are with the University of Toronto Robotics Institute, Canada (corresponding e-mail: {\tt\small spencer.teetaert@robotics.utias.utoronto.ca})}%
\thanks{${}^2$ This author began the work at ${}^1$ and is now with the Department of Computer Science, Toronto Metropolitan University, Canada }}%
}

\onecolumn

\vspace*{2cm}

% ---- Accepted-paper notice box ----
\begin{mdframed}[
    linewidth   = 1pt,
    linecolor   = black,
    innertopmargin    = 12pt,
    innerbottommargin = 12pt,
    innerleftmargin   = 14pt,
    innerrightmargin  = 14pt,
    backgroundcolor   = gray!8
]
 
\begin{center}
  {\large\bfseries Accepted Paper}
\end{center}
 
\medskip
 
\noindent This paper has been accepted for publication in the proceedings of the \textbf{IEEE-RAS International Conference on Soft Robotics (RoboSoft \textit{2026})} and is posted here as the accepted author version in accordance with IEEE's preprint posting policy.
 
\medskip
 
\noindent \textbf{Full citation and DOI:}
\hspace*{1em}
\noindent \href{https://doi.org/10.1109/RoboSoft67810.2026.11522922}{https://doi.org/10.1109/RoboSoft67810.2026.11522922}
 
\medskip
 
\noindent The final published version of record is available on IEEE Xplore at the DOI link above.  This version may differ from the published version in formatting and minor editorial details.  No other changes have been made to the accepted manuscript.

\end{mdframed}
 
\vspace{1.5cm}
 
% ---- IEEE copyright statement (required verbatim by IEEE policy) ----
\begin{mdframed}[
    linewidth   = 0.6pt,
    linecolor   = black!50,
    innertopmargin    = 10pt,
    innerbottommargin = 10pt,
    innerleftmargin   = 14pt,
    innerrightmargin  = 14pt
]
 
\small \textcopyright\ 2026 IEEE. Personal use of this material is permitted. Permission from IEEE must be obtained for all other uses, in any current or future media, including reprinting/republishing this material for advertising or promotional purposes, creating new collective works, for resale or redistribution to servers or lists, or reuse of any copyrighted component of this work in other works.
 
\end{mdframed}
 
\vfill
 
\begin{center}
  \footnotesize
  \textit{This page is required by IEEE policy for accepted papers posted publicly. The technical content of the paper begins on the following page.}
\end{center}

\twocolumn

\newpage   % <-- paper content begins after this page

\maketitle
\thispagestyle{empty}
\pagestyle{empty}

\begin{abstract}
  Stochastic state estimation methods for continuum robots (CRs) often struggle to balance accuracy and computational efficiency. While several recent works have explored sliding-window formulations for CRs, these methods are limited to simplified, discrete-time approximations and do not provide stochastic representations. In contrast, current stochastic filter methods must run at the speed of measurements, limiting their full potential. Recent works in continuous-time estimation techniques for CRs show a principled approach to addressing this runtime constraint, but are currently restricted to offline operation. In this work, we present a sliding-window filter (SWF) for continuous-time state estimation of CRs that improves upon the accuracy of a filter approach while enabling continuous-time methods to operate online, all while running at faster-than-real-time speeds. This represents the first stochastic SWF specifically designed for CRs, providing a promising direction for future research in this area. 
\end{abstract}

\section{Introduction}

Continuum robots (CRs) are flexible, small-scale manipulators capable of bending into highly nonlinear shapes and adhere to complex trajectories in confined spaces.
This allows them to operate in environments traditional rigid-link robots typically cannot enter, making them suitable for a number of previously inaccessible applications.
Examples include minimally invasive surgery~\cite{Burgner-Kahrs2015}, industrial inspection and repair~\cite{Dong2017}, and search-and-rescue in disaster areas~\cite{Hawkes2017}.

Controlling CRs in such  applications requires accurate localization within their environment.
Significant progress has been made in the physical modeling of CRs, predicting their resulting shape given actuation inputs and interaction forces~\cite{Armanini2023}.
Nevertheless, such open-loop methods still suffer from inaccuracies arising from unmodeled effects, approximated material properties, and unknown disturbances or external forces.
Consequently, integrated sensing becomes crucial to compensate for these inaccuracies and probabilistic approaches that fuse noisy measurements with suitable prior models have received increased attention.

\begin{figure}[!tp]
	\centering
	\includegraphics[width=\columnwidth]{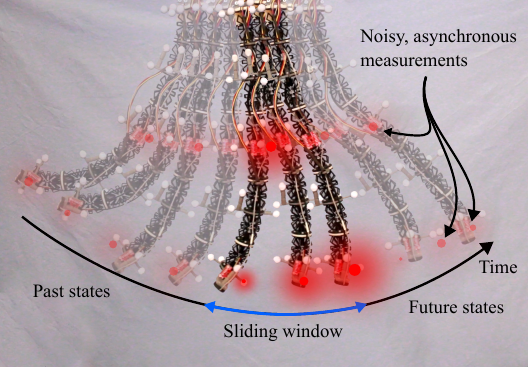}
	\caption{An example CR state estimation scenario where sensor measurements are produced asynchronously in time. The SWF in this work uses a small window of time to jointly optimize states within the window, improving accuracy over filter-based approaches while enabling online operation, something previous batch methods cannot do.}
	\label{fig:title}
\end{figure}

The vast majority of existing work on probabilistic state estimation employs filtering methods, such as extended Kalman filters (EKF) or particle filters.
These methods are computationally efficient and recursive, exploiting the Markov property, which implies that future states only depend on the current state and not the full history~\cite{Barfoot2024}.
This makes them well suited for online and real-time control scenarios, where estimates must be updated sequentially as new data arrives.
However, during the derivation and implementation of most filters, approximations are introduced, e.g., linearizations, such that the approximated posterior may no longer fully capture the Markov structure.
This can lead to biased, suboptimal, and consequently inaccurate posterior estimates for nonlinear systems.

Batch estimation methods, such as smoothers, offer a solution to this problem, as they jointly optimize all states using available measurements, retrospectively correcting past approximation errors and producing more accurate posterior estimates.
However, such approaches are typically not applicable in online settings, since only past measurements are available at any given time step.
Sliding-window filters provide a compromise between filtering and full batch optimization.
They perform smoothing over a fixed-length time window, improving estimation accuracy while remaining computationally efficient.
To date, such estimators remain largely unexplored in continuum robotics.

This paper directly addresses this gap in the literature and proposes a probabilistic sliding-window filtering approach for CRs.
The method is derived from recent batch optimization techniques~\cite{Teetaert2025} and enables online estimation of CR states.
It is validated on a variety of trajectories using a real-robot prototype, demonstrating improved accuracy over a filtering approach while maintaining real-time operation.
The estimation accuracy is shown to be comparable to full batch methods \edit{when applied to the robot shown in Fig.~\ref{fig:title}.}
To the best of the authors’ knowledge, this work represents the first probabilistic sliding-window filtering approach specifically designed for CRs.
An open-source implementation of the approach is made available to the community at \edit{\url{https://github.com/utiasASRL/space_time_continuum}.}

\section{Related Work}

Most state estimation methods for CRs adopt simplified, non-probabilistic shape representations, such as constant-curvature~\cite{Roesthuis2014,Stella2023} or polynomial-curvature models~\cite{Kim2014,Pei2025}.
Recent work fuses highly detailed dynamic Cosserat rod models with sensed data to capture fully continuous task-space states~\cite{Zheng2023,Zheng2024,Zheng2025}, but these methods remain non-stochastic and cannot account for sensor noise or quantify uncertainty.

Probabilistic approaches typically track simplified models over time using filtering methods.
Particle filters have been proposed to track constant-curvature representations of catheters over time~\cite{Brij2010,Borgstadt2015}.
Similarly, various Kalman filter implementations have been proposed to fuse noisy measurements with constant-curvature representations of CRs.
For example, extended Kalman filters have been applied to tendon-driven CRs in~\cite{Ataka2016,Peng2024}.
The state estimation of a multi-backbone CR using an unscented Kalman filter is presented in~\cite{Chen2019a}.
Filtering approaches for pneumatically actuated soft robots include both extended Kalman filters~\cite{Loo2019,Kim2021} and unscented Kalman filters~\cite{Mehl2024}.

Beyond filtering, probabilistic smoothing has been applied to state estimation problems for CRs.
Early work employs Rauch-Tung-Striebel (RTS) smoothing applied along the robot's arc length to recover quasi-static continuous shape estimates featuring variable curvature~\cite{Mahoney2016a,Anderson2017}.
However, these approaches lack temporal smoothing, as they only consider the robot’s shape along the spatial domain at a single instant.
Analogously, recent work introduces Gaussian process (GP) regression over CR arc length to estimate quasi-static states~\cite{Lilge2022,Lilge2025a}.
Lately, these approaches have been extended to include both temporal and spatial smoothing within a batch optimization framework~\cite{Teetaert2024,Teetaert2025}.

As discussed earlier, temporal filtering and batch smoothing represent two extremes in estimation approaches.
Filtering methods are computationally efficient and can be run online, but they typically offer limited accuracy.
In contrast, smoothing methods provide superior estimation accuracy, yet they are not suitable for online settings and generally scale poorly.
A sliding-window approach offers a practical trade-off, aiming to combine the advantages of both methods by balancing computational efficiency and estimation accuracy.
Sliding-window filtering methods have been studied in other robotics domains, such as mobile robot simultaneous localization and mapping (SLAM)~\cite{Huang2011} or planetary surface estimation for autonomous landing~\cite{Sibley2010}, but remain underexplored in continuum robotics.

First advances toward sliding-window estimation for CRs are presented in~\cite{Abdelaziz2023} and~\cite{Bastos2024}, although both approaches exhibit certain limitations.
In~\cite{Abdelaziz2023}, a simplified constant-curvature model is used for the sliding-window implementation, which significantly restricts the range of shape deformations that can be captured.
While this approach may perform well in free-space scenarios, it is likely to fall short under environmental interactions.
In contrast,~\cite{Bastos2024} approximates the CR shape using a rigid-link model and limits the evaluation of the sliding-window estimator to simulations.
Moreover, neither approach provides a probability distribution of the estimated posterior, limiting the ability to quantify its uncertainty.

In conclusion, a gap exists in the current literature, which this paper addresses by introducing \edit{a} probabilistic sliding-window estimator for CRs, enabling smooth estimation of their variable-curvature shape over both space and time.
\section{Methodology}

\subsection{Estimation Framework}
\begin{figure*}[!tp]
    \centering
    \includegraphics[width=\textwidth]{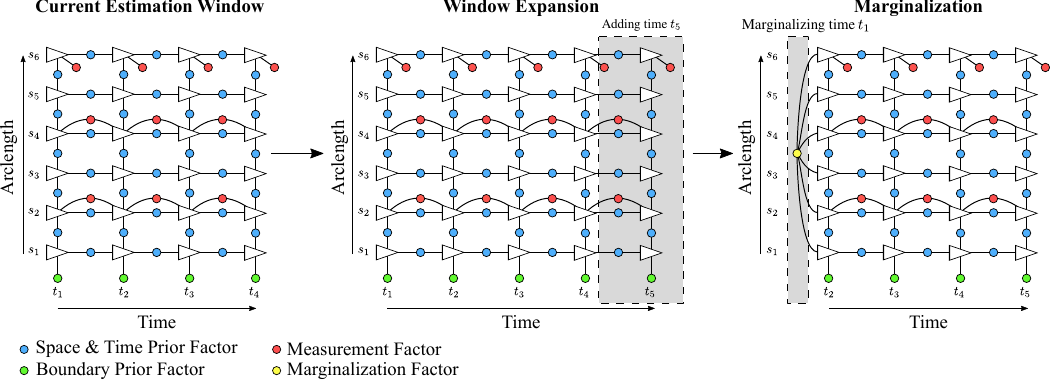}
    \caption{Factor graph representation of the sliding-window filter: The window is initially expanded to incorporate the next discrete time step. Once the new state is included, the oldest time step is marginalized out, maintaining a fixed window size while propagating information forward.}
    \label{fig:marginalization}
\end{figure*}

We construct a factor-graph estimator for estimating the state $\boldsymbol{x}$ of a continuum robot. This state includes the pose $\boldsymbol{T}(s, t) \in SE(3)$, velocity $\boldsymbol{\varpi}(s, t) \in \mathbb{R}^6$, and strain $\boldsymbol{\epsilon}(s, t) \in \mathbb{R}^6$ of the robot along its entire arc length $0 \leq s \leq L$ continuously in time for a given period $0 \leq t \leq T$. We make use of the prior and measurement factors introduced in previous work on batch smoothing~\cite{Teetaert2025} to model the robot and measurement relationships of the system. Structuring the estimation problem beginning from these factors will allow us to adopt the continuous-time methodology from previous work into a sliding-window filter (SWF). The factors are used to construct a maximum a posteriori (MAP) estimation problem, which can be solved using nonlinear optimization techniques. Specifically, given measurements $\boldsymbol{y}$, and control inputs $\boldsymbol{v}$, we are finding the state $\boldsymbol{x}^*$,
\begin{align}
    \boldsymbol{x}^* & =  \argmax_{\boldsymbol{x}} p(\boldsymbol{x} | \boldsymbol{y}, \boldsymbol{v})                \\
                     & \equiv \argmin_{\boldsymbol{x}} -\sum_i \log(\phi_i(\boldsymbol{x})).
\end{align}
Each factor $\phi_i(\boldsymbol{x})$ is of the form
\begin{align}
    \phi_i(\boldsymbol{x}) & \propto
    \exp\left(-\frac{1}{2}{\boldsymbol{e}_i(\boldsymbol{x})}^T
    \boldsymbol{\Sigma}_i^{-1}\boldsymbol{e}_i(\boldsymbol{x})\right),
\end{align}
where $\boldsymbol{e}_i$ represents an error term and $\boldsymbol{\Sigma}_i^{-1}$ represent an inverse covariance for some factor $\phi_i(\boldsymbol{x})$. The prior factors used are formulated using an approximate Cosserat rod model, derived from a `white-noise-on-acceleration' motion prior shown in past works to be effective for both mobile robotics~\cite{Anderson2015} and continuum robotics~\cite{Lilge2022, Lilge2025a, Teetaert2025}. In practice, this means our control inputs $\boldsymbol{v}$ are zero, as the prior factors encapsulate the dynamics of the system. The measurement factors used include tip pose measurements and gyroscope measurements. This optimization problem is solved by linearizing each cost term into a linear least-squares problem and iterating using the Gauss-Newton method until convergence. Specifically,
\begin{align}
    \boldsymbol{H}^T \boldsymbol{W}^{-1} \boldsymbol{H} \delta \boldsymbol{x}
     & = -\boldsymbol{H}^T \boldsymbol{W}^{-1} \boldsymbol{e}(\boldsymbol{x})
\end{align} is iterated until convergence, where $\boldsymbol{H}$ is the Jacobian of stacked errors $\boldsymbol{e}(\boldsymbol{x})$, and $\boldsymbol{W}$ is the block diagonal matrix of stacked covariances $\boldsymbol{\Sigma}_i$. Upon convergence, we extract the state covariance using the Laplace approximation as
\begin{align}
    \boldsymbol{\Sigma} & = {\left( \boldsymbol{H}^T \boldsymbol{W}^{-1}
    \boldsymbol{H} \right)}^{-1}.
\end{align}
We refer the reader to~\cite{Teetaert2025} for additional details and a more in-depth introduction of the estimation framework.

\subsection{Sliding-Window Filter Formulation}
Let us define a window with $w$ time steps that include states
$\boldsymbol{x}_{a:b}$, with $b = a + w$. We assume that the window itself has
the Markov property and is only dependent on the preceding state and window
measurements. The window contains several classes of factors:
\begin{align}
    \nonumber \phi_p(\boldsymbol{x}_{a-1})                        & \quad\text{prior factors preceding
    the window,}                                                                                        \\
    \nonumber \phi_m(\boldsymbol{x}_{i-1}^{j}, \boldsymbol{x}_i^{j})
                                                                  & \quad\text{motion factors,}         \\
    \nonumber \phi_s(\boldsymbol{x}_i^{j-1},\boldsymbol{x}_i^{j}) & \quad
    \text{spatial factors,}                                                                             \\
    \nonumber \phi_b(\boldsymbol{x}_i^{0})                        & \quad\text{boundary prior factors,}
    \\
    \phi_y(\boldsymbol{x}_{i-1}^{j},\boldsymbol{x}_i^{j})                                  & \quad\text{time-interpolated measurement factors.}
\end{align}
For brevity, we will denote all spatial factors for the entire robot state at time $i$ as $\phi_s(\boldsymbol{x}_i) = \prod_{j=1}^{N} \phi_s(\boldsymbol{x}_i^{j-1}, \boldsymbol{x}_i^j)$. Similarly, $\phi_m(\boldsymbol{x}_i)$, \edit{$\phi_b(\boldsymbol{x}_{i-1}, \boldsymbol{x}_i)$} and $\phi_y(\boldsymbol{x}_i)$ are defined. The joint factorization over the sliding-window is then
\begin{align}
    \nonumber p(\boldsymbol{x}_{a:k} | & \boldsymbol{y}_{1:k}, \boldsymbol{x}_{a-1}) \;\propto\; \\
    \nonumber & \phi_p(\boldsymbol{x}_{a-1}) \phi_b(\boldsymbol{x}_{a-1}^{0}) \phi_s(\boldsymbol{x}_{a-1}) \\ % prior moving into winodow 
    \nonumber & \times \prod_{i=a}^{k} \phi_b(\boldsymbol{x}_i^{0}) \phi_m(\boldsymbol{x}_{i-1}, \boldsymbol{x}_i) \phi_s(\boldsymbol{x}_i) \phi_y(\boldsymbol{x}_{i-1}, \boldsymbol{x}_i),
\end{align}
where $p(\boldsymbol{x}_{a:k} | \boldsymbol{y}_{1:k}, \boldsymbol{x}_{a-1})$ is the posterior distribution over the window states given all measurements in the window and the preceding state $\boldsymbol{x}_{a-1}$. Since $\boldsymbol{x}_{a-1}$ lies outside the window, we marginalize it to form an effective prior factor on the first in-window state:
\begin{align}
    \nonumber \psi_a(\boldsymbol{x}_a) \;\coloneqq\; &\int \phi_m(\boldsymbol{x}_{a-1}, \boldsymbol{x}_a) \, \phi_p(\boldsymbol{x}_{a-1}) \phi_b(\boldsymbol{x}_{a-1}^{0}) \\
    & \smallshift \times \phi_s(\boldsymbol{x}_{a-1}) \phi_y(\boldsymbol{x}_{a-1}, \boldsymbol{x}_{a}) \, d\boldsymbol{x}_{a-1}.
\end{align}
This marginalized term $\psi_a(\boldsymbol{x}_a)$ acts as a single prior factor that encapsulates all information from before the active window, allowing the optimization to depend only on variables within $\boldsymbol{x}_{a:k}$. Substituting this factor back into the joint expression yields
\begin{align}
    \nonumber p(\boldsymbol{x}_{a:k} | & \boldsymbol{y}_{1:k}, \boldsymbol{x}_{a-1}) \;\propto\; \\
    \nonumber & \psi_a(\boldsymbol{x}_a) \Bigg( \prod_{i=a}^{k} \phi_b(\boldsymbol{x}_i^{0}) \phi_s(\boldsymbol{x}_i) \Bigg) \\ % prior moving into winodow 
    \label{eq:swf_factorization} & \times \Bigg(\prod_{i=a+1}^{k} \phi_m(\boldsymbol{x}_{i-1}, \boldsymbol{x}_i) \phi_y(\boldsymbol{x}_{i-1}, \boldsymbol{x}_i) \Bigg).
\end{align}
This sliding-window factor graph formulation is visually depicted in Fig.~\ref{fig:marginalization}. Note that in this form we can see three categories of factors emerge: the marginalized prior factor $\psi_a(\boldsymbol{x}_a)$, the factors that are only a function of states that vary down the arc length of the robot, and the motion factors connecting consecutive time steps within the window. The separation of all of these factors in Eq.~\eqref{eq:swf_factorization} highlights the conditional independence of the formulation also visible in Fig.~\ref{fig:marginalization}. That is, states at a later time step are conditionally independent of states at an earlier time step given the state at the current time step. This is the Markov property once again visible in the formulation. By iterating over a window of time steps and marginalizing out old states, we can mitigate the error introduced by the Markov assumption that is embedded in the structure of the factor graph itself.

\subsection{Filter Implementation}

\begin{algorithm}[!t]
    \edit{
    \caption{Sliding-Window Filter Algorithm}
    \label{alg:swf}
    \begin{algorithmic}[1]
        \State \textbf{Initialize:} $\psi_a(\boldsymbol{x}_a) \gets \text{Prior Information}$
        \State \textbf{Initialize:} Window states $\mathcal{X} \gets \{ \boldsymbol{x}_a \}$
        \State \textbf{Initialize:} Factor graph $\mathcal{G} \gets \{ \psi_a(\boldsymbol{x}_a) \}$
        \For{$k = a+1, \dots, T$}
        \State Receive new measurements $\boldsymbol{y}_k$
        \State Initialize new state $\boldsymbol{x}_k$
        \State Add $\boldsymbol{x}_k$ to $\mathcal{X}$
        \State Add factors to $\mathcal{G}$:
        \State \quad Motion: $\phi_m(\boldsymbol{x}_{k-1}, \boldsymbol{x}_k)$
        \State \quad Spatial: $\phi_s(\boldsymbol{x}_k)$
        \State \quad Boundary: $\phi_b(\boldsymbol{x}_k^0)$
        \State \quad Measurement: $\phi_y(\boldsymbol{x}_{k-1}, \boldsymbol{x}_k)$
        \State Solve MAP estimate: $\mathcal{X}^* = \argmax_{\mathcal{X}} \prod_{f \in \mathcal{G}} f(\mathcal{X})$
        \If{$|\mathcal{X}| > w$}
        \State Define oldest state $\boldsymbol{x}_{\text{old}} \in \mathcal{X}$
        \State Compute marginal prior: $\psi_{\text{new}}(\boldsymbol{x}_{\text{new}}) = \int \mathcal{G}(\boldsymbol{x}_{\text{old}}, \dots) \, d\boldsymbol{x}_{\text{old}}$
        \State Remove $\boldsymbol{x}_{\text{old}}$ and associated factors from $\mathcal{G}$
        \State Add $\psi_{\text{new}}(\boldsymbol{x}_{\text{new}})$ to $\mathcal{G}$
        \State Remove $\boldsymbol{x}_{\text{old}}$ from $\mathcal{X}$
        \EndIf
        \EndFor
    \end{algorithmic}}
\end{algorithm}

We now face the challenge of implementing a practical solution for the aforementioned factor graph optimization problem. \edit{Pseudocode for the proposed SWF is shown in Algorithm~\ref{alg:swf}.} During estimation, we are constantly expanding our estimation window and marginalizing out old states as we go. How we extract states and choose a window size also requires attention. The details of these four items are described below.

\subsubsection{Window Expansion}
At each new time step, we initialize $N$ new states in our factor graph, corresponding to points along the robot at the new time. They are connected to each other via `space binary factors' and to the previous time step via `time binary factors' as described in previous work~\cite{Teetaert2025}. Measurement factors are also added to the new states as appropriate. A principled way to initialize the mean of these states is to use the mean produced by taking a step with the prior model. As this setup has each new state connected to two separate factors, it is not necessarily possible to initialize the new states this way. To resolve this, we find two different initializations for the new states: one that is consistent with the prior factor in space, and one that is consistent with the prior factor in time. We then average these two initializations to find a practical compromise for the initial mean. \edit{Vector values are averaged in the usual way, while pose values are averaged using an $SE(3)$-sensitive method,
\begin{equation}
    \boldsymbol{T}_{\text{avg}} = \boldsymbol{T}_1 \exp\left(\frac{1}{2} \log\left(\boldsymbol{T}_1^{-1} \boldsymbol{T}_2\right)^\vee\right).
\end{equation}}

\subsubsection{Marginalization}
To keep the size of the window bounded, and to maintain information from before the window, we marginalize out old states. Simultaneously, these states are removed from the estimation problem, locking them in place. We start with a joint Gaussian distribution over the variable we wish to marginalize, $\boldsymbol{x}_m$, and those we wish to keep, $\boldsymbol{x}_r$,
\begin{equation}
    p\begin{pmatrix}\begin{bmatrix}
            \boldsymbol{x}_r \\ \boldsymbol{x}_m
        \end{bmatrix} \end{pmatrix} \sim \mathcal{N}\left(
    \begin{bmatrix}\boldsymbol{\mu}_r \\ \boldsymbol{\mu}_m \end{bmatrix},
    \begin{bmatrix}\boldsymbol{\Sigma}_{rr} & \boldsymbol{\Sigma}_{rm} \\
               \boldsymbol{\Sigma}_{mr} & \boldsymbol{\Sigma}_{mm}\end{bmatrix} \right).
\end{equation}
The marginal distribution over $\boldsymbol{x}_r$ is given as
\begin{equation}
    p(\boldsymbol{x}_r) \sim \mathcal{N}\left( \boldsymbol{\mu}_r,
    \boldsymbol{\Sigma}_{rr} \right).
\end{equation}
In practice, we will perform this marginalization in information form, where the information matrix $\boldsymbol{H} = \boldsymbol{\Sigma}^{-1}$ and information vector $\boldsymbol{h} = \boldsymbol{\Sigma}^{-1}\boldsymbol{\mu}$. The marginalization is then performed via the Schur complement,
\begin{align}
    \boldsymbol{H}_r & = \boldsymbol{H}_{rr} - \boldsymbol{H}_{rm}
    \boldsymbol{H}_{mm}^{-1} \bold{H}_{mr},                        \\
    \boldsymbol{h}_r & = \boldsymbol{h}_r - \boldsymbol{H}_{rm}
    \boldsymbol{H}_{mm}^{-1} \boldsymbol{h}_m.
\end{align}
The joint information matrix is constructed during optimization using the Laplace approximation, and the marginalization information is iterated with the remainder of the problem. As the window slides, the information associated with only locked states is accumulated and carried forward to future time steps without the need for iteration.

\subsubsection{State Extraction}
At each time step, we wish to extract a mean and covariance estimate. While we could extract the state mean and covariance from any point in the window, the best estimate will come from the state at the back of the window, as it has been updated with all measurements in the window. However, this introduces latency in the estimate by the period of the window. Conversely, extracting from the front of the window provides a live update at the expense of losing state updates from future measurements, reducing estimation accuracy. In this work, we present results from the former approach, as the latency introduced for all practical window choices remains small ($\leq$0.1s) for the data set used.

One quirk of the continuous-time formulation arises when considering how to extract covariance estimates from the filter. The continuous-time covariance interpolation expressions in~\cite{Teetaert2025} depend on the joint covariance between each neighboring state in time. As we lock states and slide the window, the joint covariance estimates are no longer consistent between window locations as the states are relinearized. To address this, we store an estimate for the joint covariance between each pair of neighboring times as they are locked and use these during interpolation. This results in having two sources of variance estimates for the robot at each time step, which do not necessarily align. In practice, this discrepancy is typically small, but it can result in some discontinuities in the covariance estimate over time.

\subsubsection{Window Size}
In this formulation, the choice of window size is critical. It should be clear that setting the window size to include all time steps results in the original batch method. Conversely, setting the window size to only include a single time step results in a filter method, which, given the setup in this paper, results in an iterated filter algorithm. One can strike a balance between accuracy and computational efficiency by selecting an appropriate window size for the given application. In the filter case, only one time step is explicitly represented in the state. As such, the covariance extracted at each time step only reflects uncertainty at that time step (as opposed to a joint between two times), and does not support continuous-time querying of the state covariance after the fact. In contrast, larger window sizes enable after-the-fact continuous-time interpolation. In both cases, measurements within the window/step are still used in continuous measurement factors during estimation.

\subsection{Experimental Setup}

\subsubsection{Estimator Configurations}
We use the \edit{original} batch method as a baseline and an estimate for the upper-bound on achievable accuracy through a filter-based approach. The baseline is compared against the proposed SWF with sizes varying from 0 seconds (e.g.~filter) to 0.2 seconds. For each estimator, the robot is discretized into $N=5$ states along its length, and time steps between estimation nodes are set at a frequency of 30Hz. \edit{This discretization is chosen as it provides a reasonable trade-off between accuracy and computational efficiency. Increasing $N$ does not have a significant impact on accuracy but increases computational cost considerably.} The number of temporal states included in the estimator, $K$, varies with window size. The noise parameters for the factors in each estimator are kept consistent across all methods for fair comparison. All experiments are run on an Intel i7-13850HX CPU @ 3.80 GHz with 64GB of RAM.

\subsubsection{Dataset}\label{sec:experiments}
We evaluate each method on \edit{a} dataset collected on the 3D printed tendon-driven continuum robot shown in Fig.~\ref{fig:title}~\cite{Dewi2024}. The dataset consists of five different trajectories, each 10 seconds long, with varying motion characteristics and contact interactions. The individual trajectories are labelled as ``Out-of-Bounds'', ``Fast Contact'', ``Impulse 1'', ``Impulse 2'', and ``Slow Free Space''. A full description of each trajectory can be found in~\cite{Teetaert2025}. The robot is 46.6cm long and has an outer diameter of 3.6cm. It is resistant to elongation and shear and can be reasonably approximated by a Kirchhoff rod, though we note we do not constrain the estimator to such. The robot is equipped with two 6-degree-of-freedom (DoF) electromagnetic pose sensors (Aurora v3, Northern Digital Inc., Canada) at its tip and base, and two gyroscope units (ISM330DHCX, STMicroelectronics NV, Netherlands) mounted at the midpoint and tip of the robot. Ground truth pose measurements are collected at five points along the robot using an external motion capture system (Vicon Motion Systems Ltd., UK). All data is collected asynchronously, resulting in varying timestamps for each of the sensor modalities.

\subsection{Metrics}
We evaluate each method using the root mean square error (RMSE) between estimated and ground truth tip poses. For position values, we normalize the result with respect to the length of the robot. As the estimated nodes and the ground truth measurements are typically asynchronous, all metrics are evaluated using the continuous-time interpolation method provided by the estimation framework. The average normalized estimation error squared (NEES) is computed to evaluate the consistency of each estimator. As the filter method does not support continuous-time interpolation of covariances, presenting the NEES is not possible. Finally, the average runtime per time step is measured for each method to assess computational efficiency.
\section{Results}\label{sec:results}

\begin{table}[!tp]
    \centering
    \caption{Proposed estimator and baseline performance summary}
    \setlength{\tabcolsep}{0pt}
    \begin{tabular*}{\columnwidth}{@{\extracolsep{\fill}}lcccc} \toprule
        \multirow{2}{*}{\textbf{Trial}} & \multicolumn{2}{c}{\textbf{Tip RMSE}} & \multirow{2}{*}{\textbf{Avg. NEES}} & \textbf{Avg. Runtime} \\
         & Pos (\%) & Rot (rad) & & (ms) \\ \midrule
        \textbf{Out-of-Bounds} & & & & \\
        Filter (0s) & 2.38 & 0.045 & N/A & 4.4 \\
        SWF (0.1s) & 1.88 & 0.042 & 6.93 & 10.2 \\
        Batch (10s) & 2.04 & 0.041 & 7.60 & 1359 \\ \midrule
        \textbf{Fast Contact} & & & & \\
        Filter (0s) & 1.75 & 0.048 & N/A & 4.0 \\
        SWF (0.1s) & 1.29 & 0.038 & 3.56 & 10.2 \\
        Batch (10s) & 1.29 & 0.038 & 3.80 & 1379 \\ \midrule
        \textbf{Impulse 1} & & & & \\
        Filter (0s) & 1.80 & 0.053 & N/A & 3.9 \\
        SWF (0.1s) & 1.70 & 0.042 & 5.96 & 8.7 \\
        Batch (10s) & 1.70 & 0.042 & 6.54 & 1377 \\ \midrule
        \textbf{Impulse 2} & & & & \\
        Filter (0s) & 2.11 & 0.075 & N/A & 4.0 \\
        SWF (0.1s) & 1.44 & 0.050 & 4.99 & 8.9 \\
        Batch (10s) & 1.43 & 0.050 & 5.27 & 1380 \\ \midrule
        \textbf{Slow Free Space} & & & & \\
        Filter (0s) & 1.16 & 0.043 & N/A & 4.0 \\
        SWF (0.1s) & 1.16 & 0.042 & 5.26 & 9.1 \\
        Batch (10s) & 1.16 & 0.042 & 5.64 & 1380 \\ \bottomrule
    \end{tabular*}\label{tab:results}
        
        \vspace{0.5em}
        
        \begin{minipage}{\columnwidth}
            \raggedright
            \footnotesize
            For each presented metrics except for NEES, the lowest value is optimal. NEES has an optimal value of 6, the number of DoFs in the pose states being evaluated. 
        \end{minipage}
\end{table}

\begin{figure}[!tp]
    \centering
    \includegraphics[width=\columnwidth]{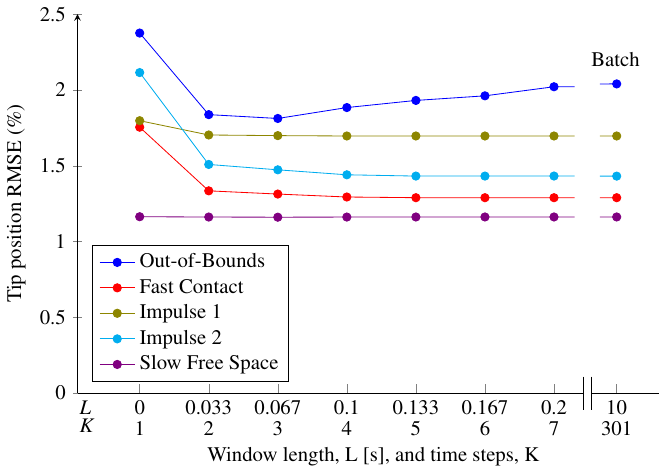}
    \includegraphics[width=\columnwidth]{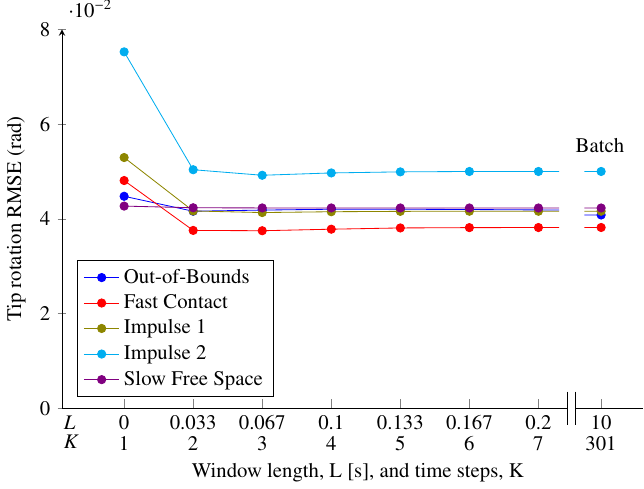}
    \caption{Tip position and rotation RMSE across each of the five experimental trajectories for different window sizes. A window size of 0s corresponds to the filter baseline, while a window size of 10s corresponds to the batch optimization baseline. In general, larger window sizes lead to lower RMSE values, with diminishing returns as the window size grows past 0.1s. Notably, even small window sizes (e.g., 0.033s) provide significant improvements over the filter baseline, indicating that incorporating a short history of states can substantially enhance estimation accuracy.}
    \label{fig:accuracy_comparison}
\end{figure}

\begin{figure*}[!tp]
    \centering
    \includegraphics[width=\textwidth]{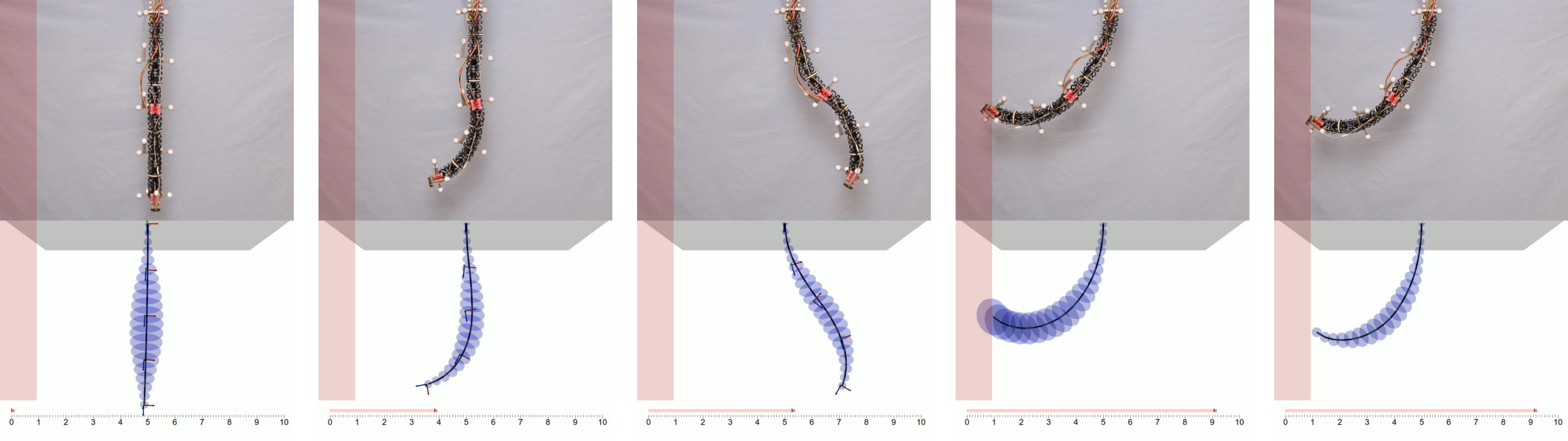}
    \caption{Side-by-side comparison of the proposed SWF with a window size of 0.1s (bottom) and the Out-of-Bounds experimental trajectory (top). The region where pose data is lost is highlighted in red on the left of the frames. The current time of each frame is provided on a visualized timeline. The SWF is able to accurately track the tip pose of the robot in real-time, even during fast motions and in the presence of occasional pose measurement dropouts. Sudden increases in uncertainty are observable when the pose measurements are lost, but the filter quickly recovers once measurements resume (see rightmost frames). }
    \label{fig:side_by_side}
\end{figure*}

\begin{figure*}[!tp]
    \centering
    \begin{minipage}{0.87\textwidth}
        \includegraphics[width=\textwidth]{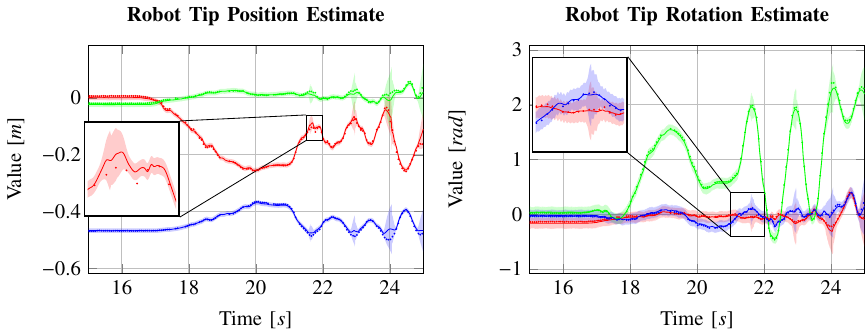}
    \end{minipage}%
    \hfill
    \begin{minipage}{0.1\textwidth}
        \includegraphics[width=\textwidth]{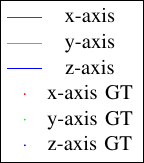}
    \end{minipage}
    \caption{Tip estimation results of a SWF with a window size of 0.1s on the Out-of-Bounds trajectory. The mean, 3$\sigma$ uncertainty bounds, and ground truth values collected from the motion capture system are shown for both position and orientation. The SWF demonstrates comparable performance to the batch optimization method, though has less smooth estimates. Notably, when the pose measurements drop out at 21.5s, the estimate contains a sudden increase in uncertainty and displays the expected discontinuity in uncertainty growth when the next measurement is recieved.}
    \label{fig:qualitative_results}
\end{figure*}
\begin{figure}[!tp]
    \centering
    \includegraphics[width=\columnwidth]{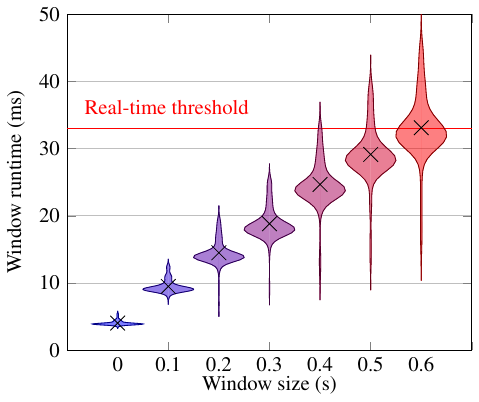}
    \caption{Window runtime distributions for different window sizes. Larger window sizes lead to higher runtimes due to their larger state size. The long tails on each distribution corresponds to different numbers of iterations required for convergence. Each window's state size is selected such that the estimated nodes are generated at a frequency of 30 Hz, as such, each window should ideally have a runtime below 33.3 ms to maintain real-time performance.}
    \label{fig:window_runtime}
\end{figure}

To select the optimal sliding-window size for our estimator, we evaluate its performance across the five experimental trajectories using various window sizes ranging from 0s (the filter baseline) to 10s (equivalent to the batch optimization baseline). The results, summarized in Fig.~\ref{fig:accuracy_comparison}, indicate that increasing the window size generally leads to improved tip position and rotation RMSE. However, the improvements exhibit diminishing returns beyond a window size of approximately 0.1s. Notably, even small window sizes (e.g., 0.033s) yield significant enhancements over the filter baseline. During the Out-of-Bounds trajectory, we do observe an increase in position RMSE when increasing the window size beyond a certain point, which is unexpected. Theories for this outlier behavior are discussed in the following section. Based on these findings, we select a window size of 0.1s to present as our primary result in the remainder of this section. 

We present qualitative results of the proposed SWF on the Out-of-Bounds trajectory. Fig.~\ref{fig:side_by_side} provides a side-by-side visual comparison of the SWF's estimated robot shape against the actual robot while Fig.~\ref{fig:qualitative_results} shows the tip pose estimates and uncertainties compared to the ground truth measurements. \edit{The periodic sudden increase in estimate uncertainty corresponds to the pose sensor's dropout.} See the 
supplementary video for a full demonstration of the SWF on the evaluated dataset. 

The proposed estimator and each baseline are evaluated on all five experimental trajectories. Their evaluation metrics are presented in Table~\ref{tab:results}. 

Lastly, for several window sizes, we evaluate the runtime of the proposed SWF over all estimation windows. We include the runtimes for each window estimated across all five trajectories. Window sizes with average runtimes that are real-time capable (i.e., below 33.3 ms) are shown in Fig.~\ref{fig:window_runtime}. The distributions exhibit multiple modes, stemming from the varying number of iterations required for convergence in each window (mostly 3, sometimes 4 or 2). These results will vary given the nature of the trajectory, as more complex motions may require more iterations to converge. We find that for our system, window sizes up to 0.3s consistently achieve real-time performance throughout operation.  
\section{Discussion}

The results clearly demonstrate the usefulness of this filter, achieving comparable accuracy to the batch solver while operating online. Even with small window sizes, significant improvements can be seen in the accuracy of the estimator. The runtimes achieved in both Fig.~\ref{fig:window_runtime} and Table~\ref{tab:results} show that even while keeping a complex, continuous shape estimate, continuum robot state estimation in real time is possible with compute to spare. This additional compute will be useful for downstream applications, such as controllers, planners, and any other processing an end user may need to perform.

Our method maintains the continuous-time properties of the original batch framework at the expense of smoothness in the estimate. Measurements can still be fed into the framework asynchronously, removing the need for any sort of pre-integration or running at extremely high frequencies. If we are comfortable losing a continuous-time covariance representation, a byproduct of this proposed SWF is a version of the filter algorithm applied to CR state estimation that runs in under 5ms (over 200Hz). This filter still provides access to a continuous representation of the state mean, which is not dependent on the joint covariance between times. We hope this formulation will enable state estimation on more dynamic systems than what has traditionally been studied in the field. 

In Fig.~\ref{fig:accuracy_comparison} we see for four of the five trajectories, the behavior of the window filter is as expected, with larger window sizes leading to better accuracy. However, in the Out-of-Bounds trajectory we see an unexpected increase in position RMSE when increasing the window size beyond a certain point. Our best hypothesis for this behavior is that when the pose sensor leaves the workspace, the measurements near the boundaries severely degrade. This could result in the batch solution overfitting to poor measurements while a shorter window filter could better handle the pose sensor dropout by relying on the gyroscopes fully. With a window method we expect the results would vary given different sensor configurations, noise profiles, and data rates. The data in this work was collected at frequencies between 30--50Hz, with low \edit{pose sensor} noise \edit{($\sim$1mm error in position, $\sim$0.01rad in rotation). We expect the proposed filter to work well with higher-noise sensors and more flexible robots, at the cost of less-certain estimates.} We \edit{also} expect that with lower frequency and higher-noise sensors, larger window sizes would be required to achieve similar performance.

The proposed method does have some drawbacks, most notably the latency introduced by extracting states from the back of the window. In practice, practitioners may wish to extract states further forward in the window based on their use case. We find that when extracting the state from the front of the window, no noticeable improvement is seen compared to the filter method. This result is surprising given the Markov property justification provided earlier. Our best hypothesis for why this occurs comes from the fact that we are not running estimation once per measurement time (as in a traditional IEKF) but rather adding multiple measurement time steps to each new window. We expect this behavior to return to expectations in cases with noisier and lower frequency data. Despite this unexpected behavior, the proposed method still provides a principled way to perform continuous-time state estimation of a continuum robot online, while maintaining uncertain estimates. \edit{Further limitations include the current assumption that all noises are zero-mean (no bias), the relatively large modelling approximations used in the prior, and the limited experimental validation on only one robot and sensor configuration. These limitations, as well as small errors in frame alignment between the sensor and motion capture frames, contribute to the observed errors in the results.}
\section{Conclusion}
The SWF proposed in this work strikes a balance between estimation accuracy, computational efficiency, and online operation capacity at the expense of introducing a small latency and reduced smoothing compared to the batch approach. As in~\cite{Teetaert2025}, we maintain a factor-graph formulation of the estimation problem, a choice we hope will lead to further adoption and a more versatile framework for others down the line. An open-source implementation for this estimator is provided for the community at \edit{\url{https://github.com/utiasASRL/space_time_continuum}.} 

\bibliographystyle{IEEEtran}
\bibliography{IEEEabrv,refs}

\end{document}